\documentclass{article}

\usepackage{arxiv}

\usepackage[utf8]{inputenc} 
\usepackage[T1]{fontenc}    
\usepackage{hyperref}       
\usepackage{url}            
\usepackage{booktabs}       
\usepackage{amsfonts}       
\usepackage{nicefrac}       
\usepackage{microtype}      
\usepackage{graphics}
\usepackage{graphicx}
\usepackage{times}
\usepackage{epsfig}
\usepackage{amsmath}
\usepackage{amssymb}

\title{Prediction of Clinical Tremor Severity Using Rank Consistent Ordinal Regression}

\author{
 Li Zhang \\
  AiCure\\
  New York, NY 10010\\
   \And
 Vijay~Yadav \thanks{corresponding author}\\
  AiCure\\
  New York, NY 10010\\
  \texttt{vijay.yadav@aicure.com} \\
    \And
 Vidya Koesmahargyo \\
  AiCure\\
  New York, NY 10010\\
  \texttt{vidya.koesmahargyo@aicure.com} \\
   \And
 Anzar Abbas \\
  AiCure\\
  New York, NY 10010\\
  \texttt{anzar.abbas@aicure.com} \\
 \And
 Isaac Galatzer-Levy \\
  AiCure\\
  New York, NY 10010\\
  }
\begin{document}
\maketitle

\begin{abstract}
Tremor is a key diagnostic feature of Parkinson's Disease (PD), Essential Tremor (ET), and other central nervous system (CNS) disorders. Clinicians or trained raters assess tremor severity with TETRAS scores by observing patients, which requires clinic visits. Lacking quantitative measures, inter or intra- observer variabilities are almost inevitable as the distinction between adjacent tremor scores is subtle. Moreover, clinician assessments also require patient visits, which limits the frequency of disease progress evaluation. Therefore it is beneficial for an automated assessment that can be performed remotely and repeatably at patients' convenience for continuous monitoring. In this work, we proposed to train a deep neural network (DNN) with rank-consistent ordinal regression using 276 clinical videos from 36 essential tremor patients. The videos are coupled with clinician assessed TETRAS scores, which are used as ground truth labels to train the DNN. To tackle the challenge of limited training data, optical flows are used as inputs to eliminate irrelevant background and statistic objects from RGB frames, only keep the crucial tremor movement patterns. In addition to optical flows, transfer learning is also applied to leverage pre-trained network weights from a related task of tremor frequency estimate. The approach was evaluated by splitting the clinical videos into training (67\%)  and testing groups (0.33\%). The mean absolute error of testing results is $0.45$, indicating that most of the errors were from the mismatch of adjacent labels, which is expected and acceptable. In addition, the model predications also agree well with clinical ratings. The model trained with the videos from ET patients is further applied to smart phone videos collected from a PD patient who has an implanted device to turn "On" or "Off" tremor. The model outputs are consistent with the patient tremor states and video observation. The results demonstrate that our trained model can be used as a means to assess and track tremor severity as clinician observation based TETRAS rating does. 
\end{abstract}


\section{Introduction}
Clinical tremor is defined as a rhythmic, involuntary oscillatory movement of a body part. Tremor is a diagnostic feature of multiple central nervous system(CNS) disorders including Parkinson’s Disease(PD), Essential Tremor(ET), and is a side effect of multiple CNS and non-CNS disorder medication treatments. Given the need to monitor clinical tremor across diverse populations who often suffer from a debilitating illness that limits mobility, there is clear value in developing automated methods of assessment that can be conducted remotely in patients' home environments. Video-based tremor quantification methods hold significant promise as a method to quantify tremor activity as  tremor is easily detected and primarily assessed visually. 

\begin{figure*}
\begin{center}
   \includegraphics[width=6.0in]{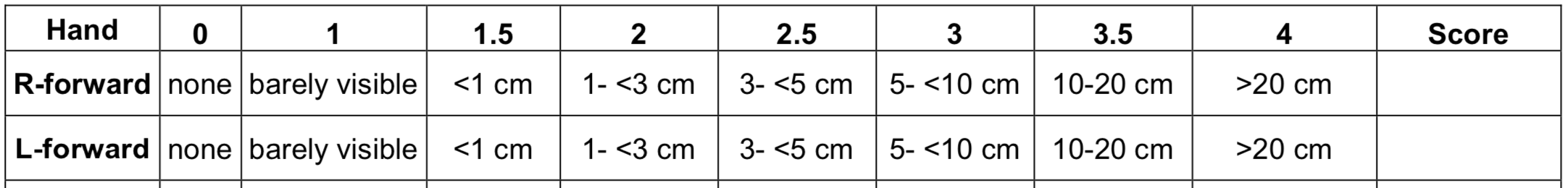}
\end{center}
   \caption{Clinical rating criteria for tremor scores. In clinical practice, a clinician or video rater fills in the score cells based on the given criteria.}
\label{fig:clinic_score}
\end{figure*}

Clinical tremor are assessed by Tremor Research Group’s Essential Tremor Rating Assessment Scale (TETRAS) contains a 9-item performance subscale that rates action tremor from 0 to 4 in half-point intervals \cite{elble2012}. Hand tremor ratings are defined by specific amplitude ranges in centimeters, see Figure \ref{fig:clinic_score}. However, as the TETRAS score is given from clinicians or raters' observation, lacking of ways to quantitatively measure movement, the reliability varies. 

To automate tremor scoring from videos, we retrieved videos of essential tremor patients as they were assessed for tremor severity. The video sessions are coupled with TETRAS score sheets for each patient. Then a deep neural network (DNN) is trained to predict TETRAS scores from videos. 

TETRAS scores are ordinal labels, which are discrete values with implicit rank information. Ordinal data is common in various applications, such as longitudinal progress predication of diseases, severity assessment of pathologies, and age estimation. While classification models are studied most extensively in machine learning and deep learning, for ordinal labels, like TETRAS scores, a typical classification model would encounter problems that samples with adjacent labels may overlap considerably and the label ordering is not retained in predication throughputs. Therefore, ordinal regression (classification), a learning method incorporates rank information intrinsically, becomes a good option for such applications. Previous research shows that ordinal regression is more accurate and robust than unordered categorical classification on large scale public datasets \cite{Frank2001OrdLabel} . Meanwhile, ordinal regression should also performs better than continuous regression as it predicts only a limited number of values, a much smaller dimensionality of output space than the infinite range of real values as in  continuous value regression, making the optimization more plausible. More recently, Niu \textit{et al.} applied ordinal regression with deep neural networks\cite{Niu2016OrdRegressCnnAge}, and Cao \textit{et al.} further improved the approach by enforcing rank consistency \cite{Cao2019coral}.  

In this work, we propose to use the rank-consistent ordinal regression \cite{Cao2019coral} to train a deep neural network (DNN) with TETRAS scores of clinical videos from ET patients (see Figure \ref{fig:workflow_rank}). In this approach, we first apply an optical flow network (FlowNet \cite{ Fischer2015flownet}) to extract object motion from consecutive frames in a video, then use transfer learning to address limited training data issue. We finally perform ordinal regression with rank consistency \cite{Cao2019coral} to train a model predicting ranked tremor score. 

\section{Method}
\begin{figure*}
\begin{center}
{\includegraphics[width=5.5in] {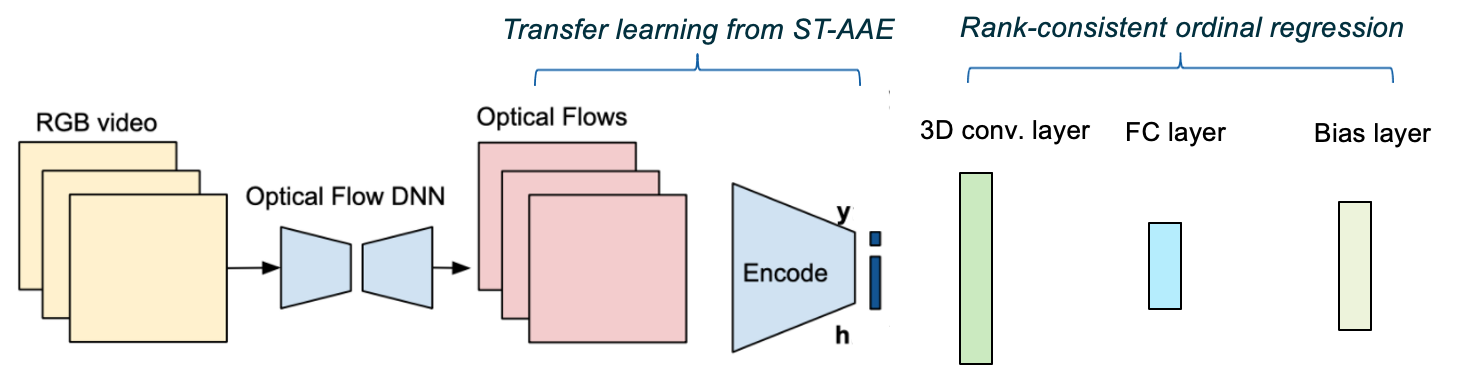}}
\end{center}
   \caption{Workflow of tremor severity prediction using rank-consistent ordinal regression with optical flow and transfer learning.}
\label{fig:workflow_rank}
\end{figure*}

As shown in Figure \ref{fig:workflow_rank}, our approach consists of three main components: optical flow calculation,  transfer learning, and rank-consistent ordinal regression. 

\subsection{Optical Flows for Tremor Movements}
\begin{figure*}
\begin{center}
{\includegraphics[width=1.0\linewidth] {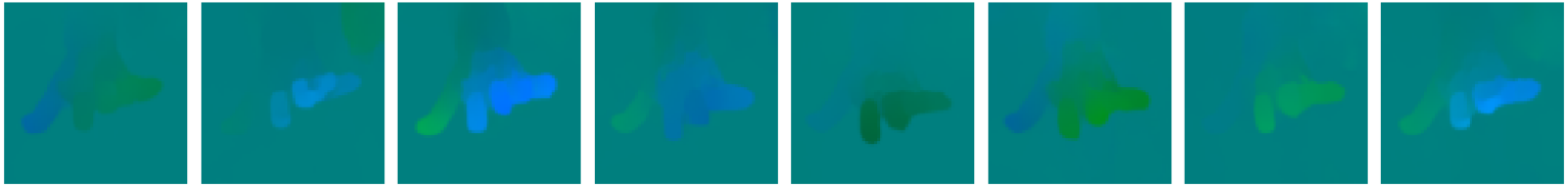}}
\end{center}
   \caption{A tremor video and its optical flows. To visualize the optical flows, the normalized $u$ and $v$ components are set as the green and blue channel value in RGB, with red channel as 0.}
\label{fig:oflow}
\end{figure*}

Optical flows \cite{Horn1981opticalflow} are widely used in video processing to represent pixel-wise change between consecutive frames. Let $\boldsymbol{p}_t$ denote a pixel at time point $t$, then its displacement from time point $t_0$ to $t_1$ defines the optical flow $(u,v) =  \boldsymbol{p}_1 - \boldsymbol{p}_0$. Figure \ref{fig:oflow} gives an example of optical flow from hand tremor. Note that the visualization of the optical flows in Figure \ref{fig:oflow} is different from the commonly used color scheme. We set the normalized $u$ and $v$ components as the green and blue channel value in RGB, with red channel as 0. 

While the combination of RGB frames and optical flows are popular in video analysis \cite{Carreira2017QuoVadis}, we only use optical flows in our proposed approach, objects (hands) themselves are not important, the characteristics of movements are. Compare with original RGB frames, optical flows from tremor videos have much less information, showing only movement patterns. That would improve model performance especially when training data is limited. More detail on the advantage of using optical flow in hand tremor analysis can be found in \cite{lizhang2020HandTremor}. 

In our work, we use fully convolutional networks, FlowNet (FlowNet 2.0) \cite{Fischer2015flownet, Ilg2017flownet2}, to infer optical flow maps from RGB videos for its reported good performance and data-driven potential attributed to deep learning. To avoid overfitting and save computational cost, we fixed the pre-trained FlowNet in this work. 

\subsubsection{Transfer Learning}
While more and more research work demonstrate the power of transfer learning, we also use transfer learning in this work to enhance the performance of our trained model, in particular, with limited training data. As in many medical applications, though the clinical videos with expert labeling we collected is from a relatively large clinical trial, it is still not sufficient to train a DNN that has a large number of parameters (weights of the neural network) and requires large amount of labeled data, typically thousands to millions of samples to train a high performance model. 

To overcome this challenge, transfer learning is exploited in model training through a previously trained tremor frequency model \cite{lizhang2020HandTremor}. The tremor frequency model is a spatio-temporal adversarial autoencoder (ST-AAE) that integrates spatial and temporal information simultaneously into the original AAE, taking optical flows as inputs. A latent space is constructed to represent the critical  features for tremor  analysis by encoding various movement patterns. From the latent space, random samples can be drawn from a prior distribution to generate synthetic image with desired labels. The ST-AAE was trained and evaluated with 3068 two-second long video segments from 28 subjects by cross-validation, and the weighted average of the AUCs of the ROCs is $0.97$. 

Since the ST-AAE model is also trained using the same optical flows as inputs, and targeted on a similar tremor analysis task, we take all the layers from the bottom to penultimate layer in the encoder of the ST-AAE model as the backbone of the DNN in this work. The last layer in the encoder of the ST-AAE is excluded as it is trained as task-specific weights for tremor frequency prediction. We do not use the decoder of the ST-AAE because it is trained to generate frequency specific tremor movement, which is irrelevant to this work.  

\subsection{Rank-Consistent Ordinal Regression}
Although ordinal labeling can simply be handled by classification models, followed by post-processing to enforce rank order, a general machine learning framework \cite{Li2007OrdRegress} with extended binary classification \cite{Frank2001OrdLabel}, is more commonly used for ordinal data. This general framework was later used together with deep neural networks \cite{Niu2016OrdRegressCnnAge, Cao2019coral}. 

\begin{figure}[t]
\begin{center}
   \includegraphics[width=0.45\linewidth]{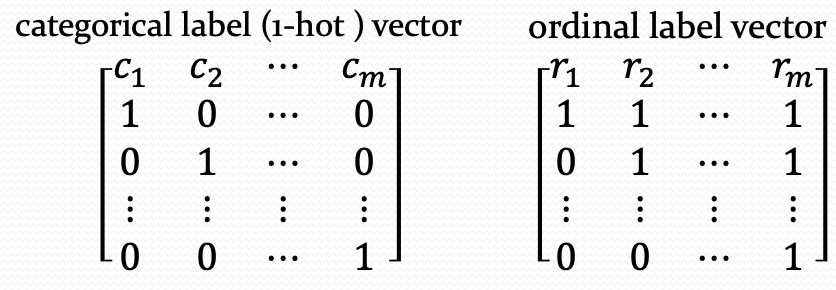}
\end{center}
   \caption{Ordinal label representation, comparing with the categorical label representation. }
\label{fig:ordinal_label}
\end{figure}

\paragraph{Ordinal label representation}
It worth to mention that ordinal regression sometimes is also referred as ordinal classification and it is more related to classification rather than regression in terms of label representation. Like classification model training, ordinal regression also uses vector representation for labeling. While classification models use one-hot vector labels, ordinal regression uses "multi-hot" vector labels. That is, if we let $r_i$ be the rank of ordinal data labels $y_i$, with the ordering $r_1 \le  r_2 \le  ... \le r_n$, then for a ranked label $r_i$, instead of only marking the corresponding vector component as $1$, ordinal regression marks $1$ for all the vector components that less than and equal to $r_i$ in order, as shown in Figure \ref{fig:ordinal_label}. 

Using the extended "multi-hot" vector representation and typical classification cross-entropy as loss function, ordinal regression penalizes more for erroneous labels farther away from the correct labels, and is more tolerate with errors adjacent to the correct labels. Let $\mathcal{L}(r_{y}, r_{c})$ be the cross-entropy loss for a sample $\boldsymbol{X}$, where $r_{c}$ is the ground labeled rank, and $r_{y}$ is the predicted rank, then $\mathcal{L}(r_{i}, r_{c}) <  \mathcal{L}(r_{j}, r_{c})$ if $|r_{i} - r_{c}| < |r_{j} - r_{c}|$. This is the desirable property we would prefer, so that the ordinal regression can be trained to embed the rank information from the labels. 

\paragraph{Rank-consistent ordinal regression}
To address the training complexity of convex cost matrix and ensure rank-monotonic threshold of binary classification extension, Cao \textit{et al.} proposed CORAL framework to product rank-consistent predication for each binary task \cite{Cao2019coral}. We follows the CORAL framework in this work, using transfer learning from our previously trained ST-AAE model. 

From the ST-AAE encoder, we first exclude the last task specific layer, then add a 3D convolutional layer for combined spatio-temporal information, followed by a fully connected layer (FC layer in Figure \ref{workflow_rank}) and a linear bias layer, as shown in Table \ref{tab:nn_layers}. 

\begin{table}
\begin{center}
\begin{tabular}{|c|c|c|c|}
\hline
No. & Layer & Parameters & Output Size \\
\hline\hline
0 &Input & & $2\times64\times64\times64$ \\
\hline
1 &Conv3d & $4\times4\times4$ $(2)$ & $64\times32\times32\times32$\\
\hline
2 &Conv3d & $4\times4\times4$ $(2)$ & $128\times16\times16\times16$\\
\hline
3 &Conv3d & $4\times4\times4$ $(2)$ & $256\times8\times8\times8$\\
 \hline
4 &Conv3d & $4\times4\times4$ $(2)$ & $512\times4\times4\times4$\\
\hline
5 &Conv3d & $4\times4\times4$ $(1)$ & $32\times1\times1\times1$\\
\hline
6 &Linear & -- & $1\times1\times1\times1$\\
\hline
7 &Bias & -- & $m - 1$ \\
\hline
8 &softmax & -- & -- \\
\hline\hline
\end{tabular}

\end{center}
\caption{The deep neural network structure in our proposed approach. Note that except the last convolutional layers, each of the 3D convolutional layers is followed by a 3D batch normalization layer and ReLU function. The Linear layer in the table is the fully connected layer. $m$ in the Bias layer is the number of labels. }
\label{tab:nn_layers}
\end{table}

%
        
\section{Results and Discussions}
\subsection{Clinical Tremor Data Acquisition}
\label{ssec:clinicdata}
We retrieved patient videos as they were assessed for tremor severity. These patients were recruited as part of a Phase 2 Clinical Trial for the drug CX-8998 conducted, registered under IRB \#201702183 \cite{ClinicData2019}. Patient criteria includes essential tremor patients between ages 18 to 75. The video sessions are coupled with TETRAS score sheets for each patient. The clinical ratings are indicated both by clinicians (in-person) and by raters from reviewing the patient videos (video). Both ratings were used for evaluation, but only clinician ratings are used for training as in general clinician ratings are more accurate and reliable than raters. 

\begin{figure}[t]
\begin{center}
   \includegraphics[width=0.45\linewidth]{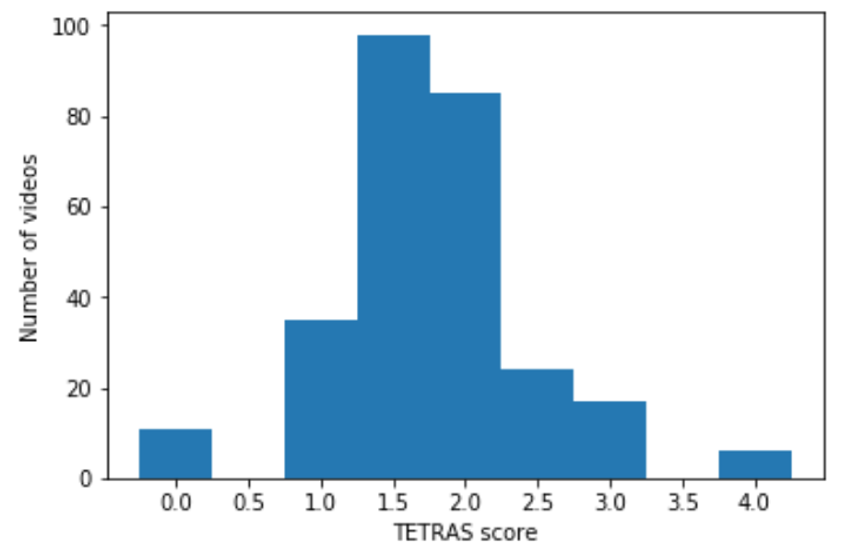}
\end{center}
   \caption{Histogram of the clinical rating scores from in-person clinician rating and remote rater rating from videos.  }
\label{fig:score_hist_all}
\end{figure}

There were in total 276 videos from 36 patients, with 1-4 video recordings for each patient, and each video recording has two separate tremor video segments for the left and right hand. Most of the patients had 4 videos that include screening, baseline, and 2 follow up videos. Each video is associated with two TETRAS scores. Figure \ref{fig:score_hist_all} shows the histogram of TETRAS scores from clinician. We can see that most of the videos were scored as mild to moderate, with very limited samples of no tremor and severe cases. With this imbalanced data distribution, our ordinal regression uses the task importance weighting from \cite{Cao2019coral}. 

To evaluate our propose approach, the clinical data is split by patients for training and testing. 186 videos from 24 patients were used for training, while 90 videos from 12 patients were kept for evaluation. 

\begin{figure}[t]
\begin{center}
   \includegraphics[width=0.6\linewidth]{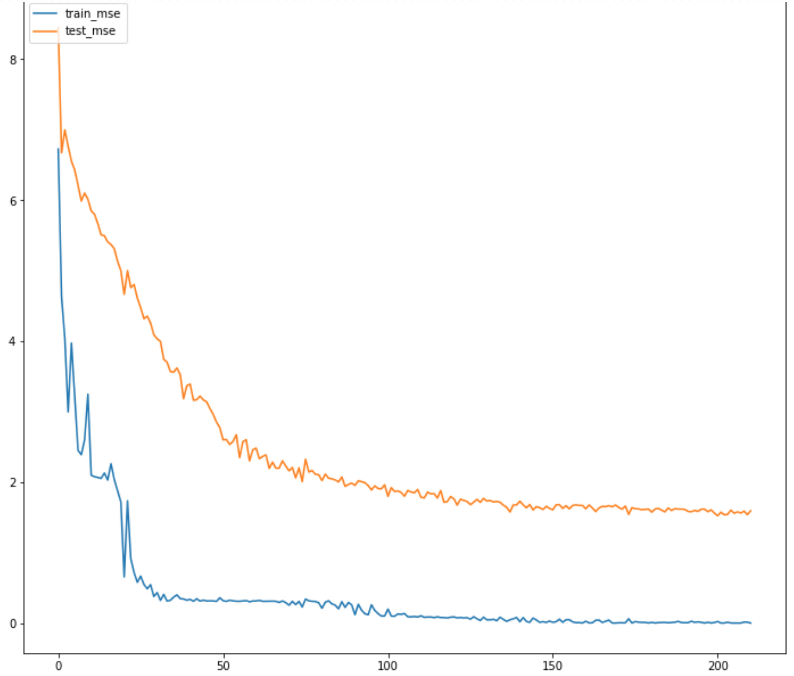}
\end{center}
   \caption{Correlation between the clinical observations and the model predictions.}
\label{fig:converge}
\end{figure}

Figure \ref{fig:converge} shows the convergence curve of training and testing mean squared error (MSE). 

The model performance was evaluated by mean absolute error and classification accuracy. While the classification accuracy is $0.36$, the mean absolute error was $0.45$.  That means most of the errors were from the mismatch of adjacent labels, which was expected and acceptable given the subtle differences among these labels and observation variabilities. 

\begin{figure}[t]
\begin{center}
   \includegraphics[width=0.75\linewidth]{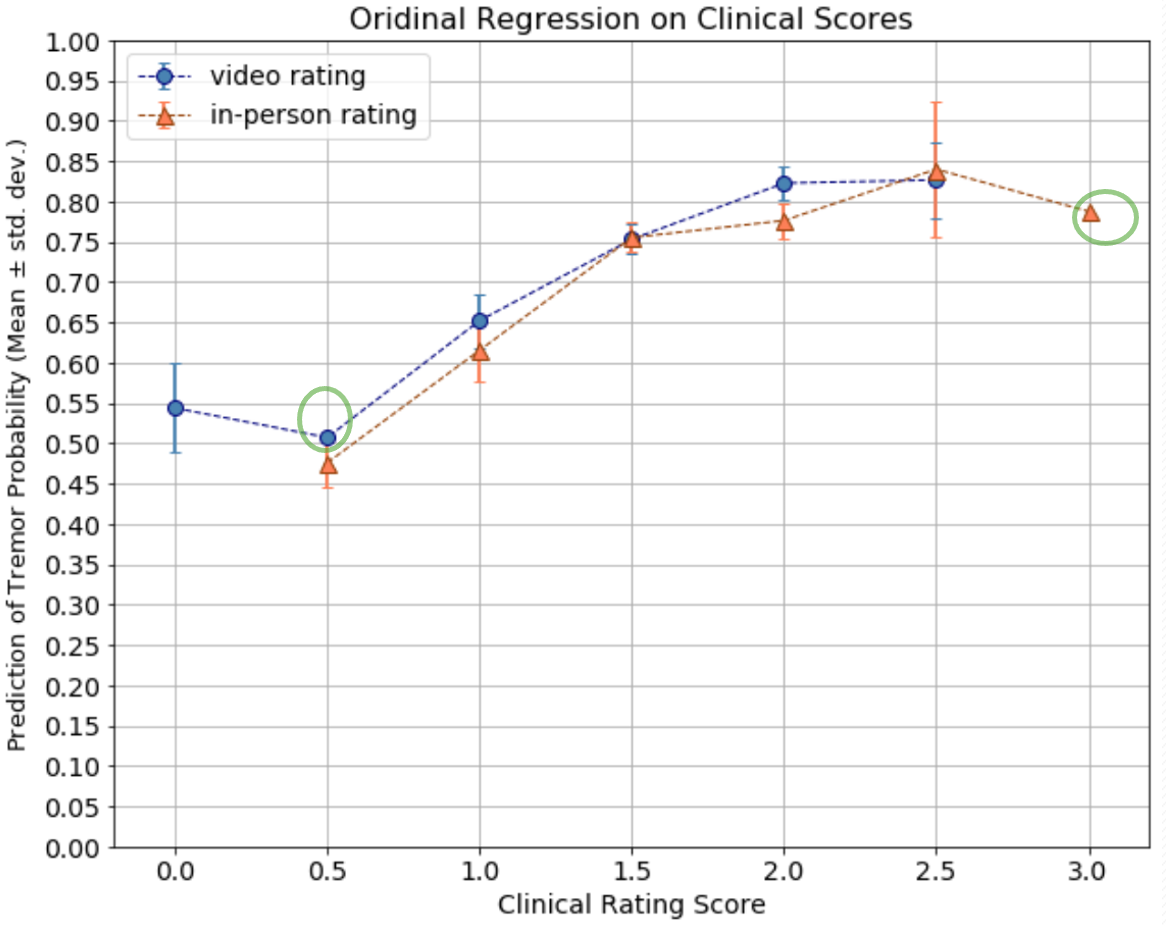}
\end{center}
   \caption{Correlation between the clinical observations and the model predictions.}
\label{fig:corr_rank}
\end{figure}

To further evaluate the efficacy of our model, we also showed the correlation of model predication on tremor probability with clinical scores, as shown in Figure \ref{fig:corr_rank}. This correlation showed the monotonic increase of tremor probabilities from model predication as TETRAS scores increase, indicating that our model can predict tremor severity for assessment as observation based TETRAS score do. The two non-monotonic points (marked by green ellipses) in Figure \ref{fig:corr_rank} were from score groups with only one video, which is rather ad-hoc and negligible. 

The model trained with essential tremor patients was also applied to videos recorded using a smart phone from a Parkinson's Disease patient. The patient has a brain implant that literally lets her turn on and off her tremor symptoms. The videos were captured when the patient symptoms were "On" or "Off". The predicted overall TETRAS score for the patient video with tremor "OFF" was zero, indicating no tremor was predicted from the model; the predicted overall TETRAS for the patient video with tremor "ON" is 2.0, indicating moderate tremor was predicted by the model (see Figure \ref{figs:pd_tremor}). The predicated overall TETRAS scores for the Parkinson's Disease patient were consistent with the tremor states. This result demonstrates the efficacy of our approach for remote assessment Parkinson's and other tremor related conditions. 

\begin{figure*}
\begin{center}
   \includegraphics[width=0.9\linewidth]{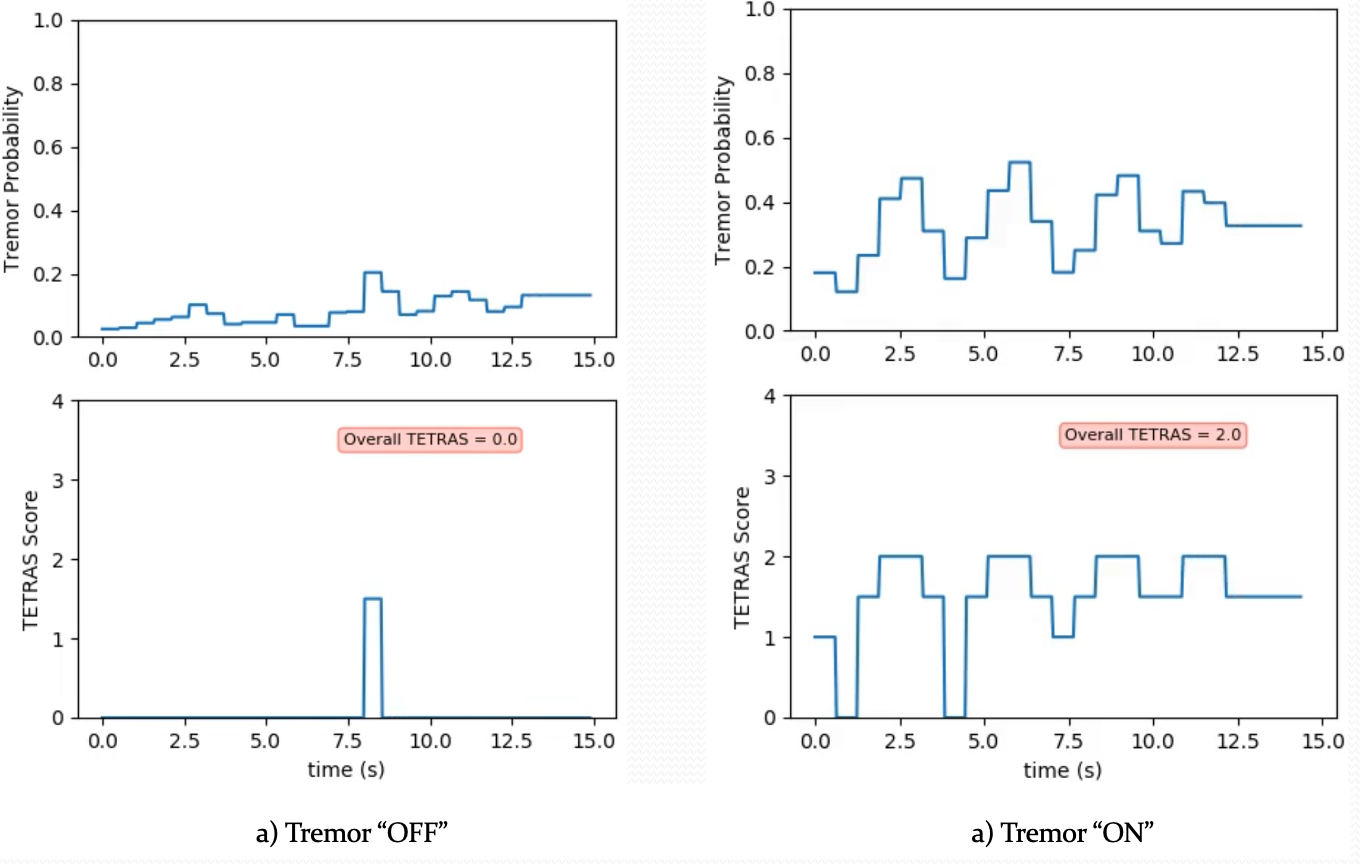}
\end{center}
   \caption{Model prediction for videos from Parkinson's Disease patient.}
\label{fig:pd_tremor}
\end{figure*}
\newpage
\bibliographystyle{unsrt}  
\bibliography{references}  

%
%
%
%

\end{document}